\pdfoutput=1
\documentclass[letterpaper, 10 pt, conference]{ieeeconf}  

\IEEEoverridecommandlockouts                              

\usepackage[OT1]{fontenc}

\usepackage{amsmath} 

\usepackage{graphicx}
\usepackage{sigmund_macros}
\usepackage{multirow}
\usepackage{soul}
\usepackage[frozencache,cachedir=.]{minted}
\usepackage{listings}
\usepackage{todonotes}
\usepackage{svg}
\usepackage{amsmath}
\usepackage{amssymb}
\usepackage{booktabs}
\usepackage{colortbl}
\usepackage{algorithm}
\usepackage{algorithmic}

\usepackage{subcaption}
\usepackage{adjustbox}

\usepackage{pgfplots}
\pgfplotsset{compat=1.17}
\usepackage{xspace}
\usepackage{url}

\usepackage{bibentry}
\definecolor{deeppink}{rgb}{1.0, 0.08, 0.58}

\usepackage{xcolor}
\usepackage{url}

\usepackage{hyperref}
\hypersetup{
  colorlinks = true,  
  urlcolor   = deeppink,  
  linkcolor  = blue,  
  citecolor  = red,  
  pdfview    = XYZ,  
}

\allowdisplaybreaks

\usepackage{hyperxmp}

\title{\LARGE \bf
Decomposed Object Manipulation via Dual-Actor Policy
}

\author{
    \textbf{Bin Fan}\textsuperscript{1}, 
    \quad \textbf{Jian-Jian Jiang}\textsuperscript{1},
    \quad \textbf{Zhuohao Li}\textsuperscript{1}, 
    \quad \textbf{Xiao-Ming Wu}\textsuperscript{2}, \\
    \quad \textbf{Yi-Xiang He}\textsuperscript{1},
    \quad \textbf{YiHan Yang}\textsuperscript{1},
    \quad \textbf{Shengbang Liu}\textsuperscript{1},
    \quad \textbf{Wei-Shi Zheng}\textsuperscript{1,3,$\dagger$} \\
    \textsuperscript{1}  \small School of Computer Science and Engineering, Sun Yat-sen University, China \\
    \textsuperscript{2}  \small College of Computing and Data Science, Nanyang Technological University, Singapore. \\
    \textsuperscript{3} \small Key Laboratory of Machine Intelligence and Advanced Computing, Ministry of Education, China \\
    {\tt\small \{fanb6, jiangjj35\}@mail2.sysu.edu.cn; wszheng@ieee.org}
}

\begin{document}
\maketitle
\pagestyle{empty}
\vspace*{-5mm}
\begin{abstract}
Object manipulation, which focuses on learning to perform tasks on similar parts across different types of objects, can be divided into an approaching stage and a manipulation stage. However, previous works often ignore this characteristic of the task and rely on a single policy to directly learn the whole process of object manipulation. To address this problem, we propose a novel Dual-Actor Policy, termed DAP, which explicitly considers different stages and leverages heterogeneous visual priors to enhance each stage. Specifically, we introduce an affordance-based actor to locate the functional part in the manipulation task, thereby improving the approaching process. Following this, we propose a motion flow-based actor to capture the movement of the component, facilitating the manipulation process. Finally, we introduce a decision maker to determine the current stage of DAP and select the corresponding actor. Moreover, existing object manipulation datasets contain few objects and lack the visual priors needed to support training. To address this, we construct a simulated dataset, the Dual-Prior Object Manipulation Dataset, which combines the two visual priors and includes seven tasks, including two challenging long-term, multi-stage tasks. Experimental results on our dataset, the RoboTwin benchmark and real-world scenarios illustrate that our method consistently outperforms the SOTA method by 5.55\%, 14.7\% and 10.4\% on average respectively.

\end{abstract}

\section{Introduction}
\label{sec:Introduction}
As a fundamental and emerging task in robotic community, object manipulation \cite{wu2025afforddp, zhu2025objectvla, Wang2025articubot, ling2024articulated} aims to perform precise and robust operations on various everyday objects that vary widely in shape, size and appearance, which has broad applications in diverse real-world scenarios, such as assembly \cite{jiang2023mastering} and household service \cite{zhang2024empowering}.

\begin{figure}
    \vspace*{-5mm}
    \centering
    \hfill
    \includegraphics[width=0.95\linewidth]{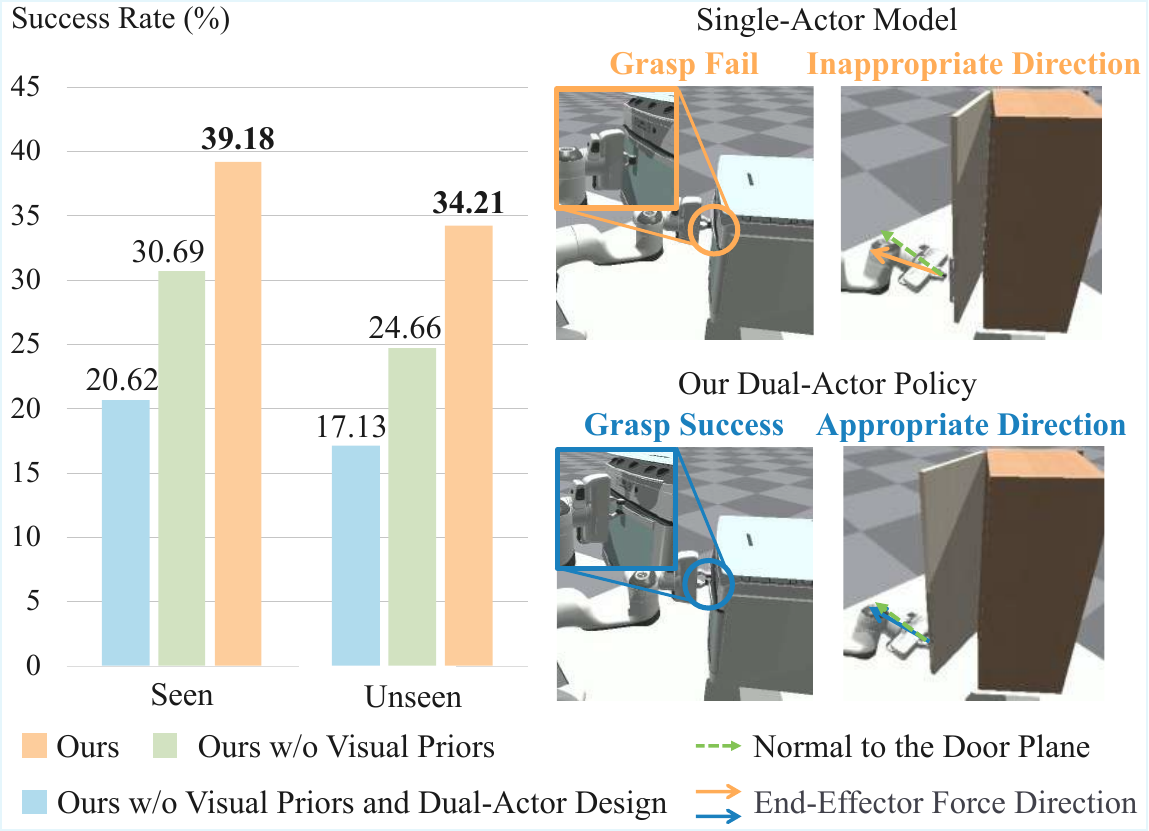}
    \caption{\textbf{Single-Actor Model vs. Our Dual-Actor Policy.} (1) Grasp fail: The single-actor model fails to grasp the handle due to inaccurate localization. (2) Inappropriate direction: The applied force direction of the single-actor model differs from the surface normal of the object components, causing detachment during door opening. Zoom in for the best view.}
    \label{fig:Single policy vs. Ours}
    \vspace{-12pt}
\end{figure}

Recently, with the rapid development of imitation learning \cite{chi2023diffusion, Ze2024DP3}, object manipulation \cite{wu2025afforddp, zhu2025objectvla, chen2025g3flow}, which can be divided into an approaching stage and a manipulation stage, achieves significant progress. However, previous works often ignore this important characteristic and typically rely on a single policy to directly learn how to manipulate objects. This sometimes results in incorrectly reaching the task-relevant components or improperly manipulating the intended parts of the objects, as illustrated in Fig. \ref{fig:Single policy vs. Ours}.

\begin{figure}
    \centering
    \hfill
    \includegraphics[width=0.95\linewidth]{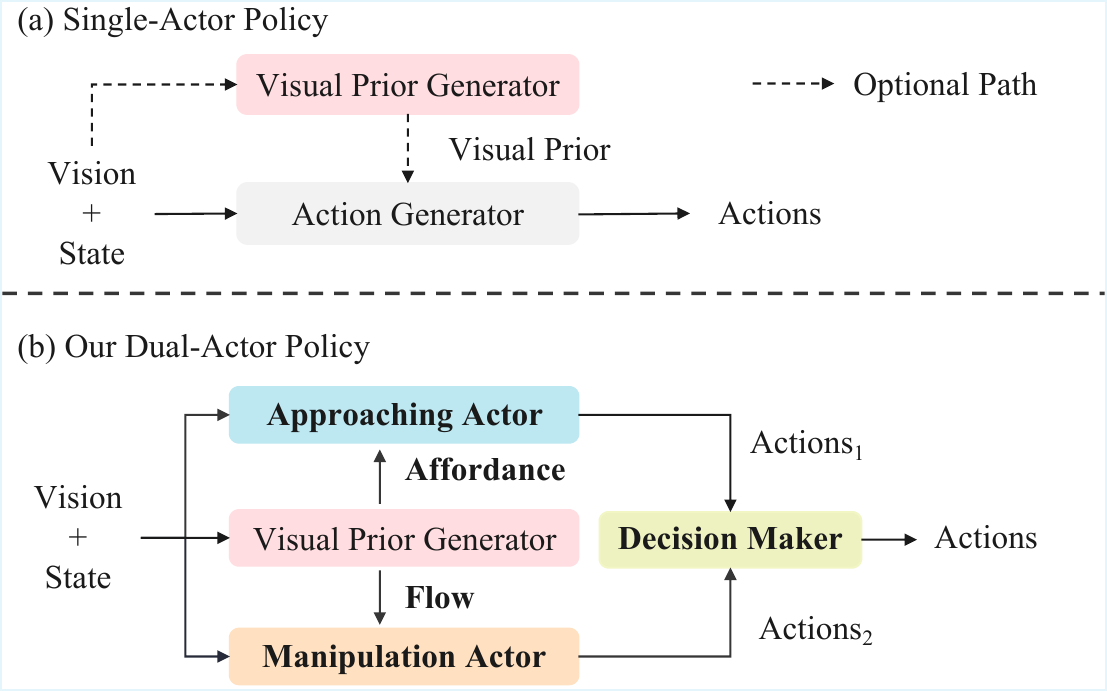}
    \caption{\textbf{Comparison between Single-Actor Models and Our Dual-Actor Policy.} Single-actor models (a) learn the entire manipulation task using a single policy, with some relying on a single type of visual prior for action prediction. In contrast, our dual-actor policy (b) decouples the object manipulation task into approaching and manipulation stages. Each stage is processed by a specialized actor, enhanced with appropriate visual priors for action prediction, while a decision maker controls the actor switching, effectively improving task performance.}
    \label{fig:Framework}
    \vspace{-12pt}
\end{figure}

Towards this end, we propose a novel stage-aware Dual-Actor Policy, termed DAP, which fully considers the characteristics of stage decoupling and utilizes heterogeneous visual priors for each stage to improve performance. Compared with previous works, as shown in Fig. \ref{fig:Framework} (a), which directly predict actions with a single policy, our framework, as shown in Fig. \ref{fig:Framework} (b), uses a dual-actor policy consisting of an affordance-based approaching actor, a motion flow-based manipulation actor and a decision maker that collaboratively predicts actions. The decision maker autonomously determines which actor to use based on the current task progress. As illustrated in Fig. \ref{fig:Single policy vs. Ours}, our method correctly reaches the specified components and properly completes tasks.

Specifically, as affordance can provide effective spatial guidance for contact and indicate where to interact, we combine affordance with a diffusion-based policy to guide precise component localization and grasping. Following that, as motion flow can capture the overall movement trends of object parts and inform the expected motion during interaction, we incorporate motion flow into a diffusion-based policy to handle the manipulation phase across diverse instances. Finally, based on our detailed phase annotations, we train the decision maker to adaptively schedule the two actors according to scene observations. With the task decomposition characteristic and multi-actor collaboration, we significantly improve robustness and generalization in complex and cross-instance manipulation scenarios.

Moreover, since previous object manipulation datasets \cite{chen2025g3flow, mo2021where2act, xu2022universal} contain few objects and lack the two visual priors important for supporting the training of our method, we construct a comprehensive simulated dataset, termed Dual-Prior Object Manipulation Dataset, which includes more objects and incorporates the two visual priors. Built upon IsaacGym \cite{makoviychuk2021isaac}, our dataset leverages the digital assets and part-level annotations from GAPartNet \cite{geng2023gapartnet} and RoboTwin \cite{chen2025robotwin}. It includes seven language-guided object manipulation tasks, each involving multiple objects from diverse categories with significant variations in appearance. The dataset also contains tasks that can be naturally decomposed into sequential two-stage phases, allowing for a detailed evaluation of phase-aware decision-making and control.

Extensive experiments on our dataset, the RoboTwin benchmark \cite{chen2025robotwin} and in real-world scenarios show that our method effectively improves manipulation performance, surpassing the SOTA method by \textbf{5.55\%}, \textbf{14.7\%} and \textbf{10.4\%} on average respectively. We will open-source our code and dataset for the development of the community.

\section{Revisiting Object Manipulation}
\label{sec:Related Work}

Object grasping \cite{DBLP:conf/icpr/JiangWCWZ24, DBLP:conf/icra/WangCLLZ25, wu2024economic, chen2024motiongrasp} is a fundamental prerequisite skill in general manipulation\cite{DBLP:journals/corr/abs-2503-09186, DBLP:conf/cvpr/ZhouLLZ21, zhou2025exploring}, focusing on the stable acquisition of physical contact with an object. In contrast, object manipulation encompasses a broader class of behaviors that involve not only grasping but also subsequent interactions such as reorienting, translating, or actuating parts of objects to achieve functional goals. The goal of object manipulation is to learn task-relevant skills—such as opening a drawer—and generalize them across different instances and categories of objects. For example, a policy trained to open a cabinet drawer should transfer effectively to drawers in desks or refrigerators, despite variations in appearance and kinematics.

And these tasks can naturally be divided into two stages: the approaching stage and the manipulation stage. However, previous research mainly treats object manipulation as a monolithic process, focusing on directly manipulating objects with a single policy, where policy modeling is typically combined with different visual priors. Such visual priors include affordances \cite{wu2025afforddp, mo2021where2act, ning2023where2explore, bahl2023affordances, kuang2025ram}, motion flow \cite{xu2022universal, eisner2022flowbot3d, xu2025flow, zhi20253dflowaction}, keypoints \cite{Wang2025articubot, sundaresan2024s, ma2024hierarchical} and semantic fields \cite{chen2025g3flow, wang2025d, ze2023gnfactor}, all of which guide policies to focus on relevant object components or task-relevant dynamics. Although these methods demonstrate how different visual priors contribute to more effective object manipulation, ignoring the task decomposition still increases learning complexity and hinders generalization.

Towards this end, we propose learning object manipulation through task decomposition into the approaching and manipulation stages, while introducing a novel Dual-Actor Policy, termed DAP, which integrates heterogeneous visual priors into these stages. The key insight is that we explicitly decouple the approaching and manipulation stages into separate, specialized policies, reflecting the natural decomposition of object manipulation tasks, with each stage guided by an appropriate visual prior. We consider that different phases benefit from different visual priors, with affordances being effective for identifying viable contact points during the approaching stage and motion flow capturing rich dynamics necessary for manipulation. By combining these heterogeneous visual priors with a stage-aware design, DAP achieves more generalizable control across objects, i.e.:
\begin{equation}
\begin{aligned}
    \text{Action}_1 &= \pi_{\text{stage1}}\bigl(Lang,\, Obs,\, State,\, \textit{Afford}\bigr), \\[2pt]
    \text{Action}_2 &= \pi_{\text{stage2}}\bigl(Lang,\, Obs,\, State,\, Flow\bigr), \\[2pt]
    \text{Action}   &= \pi_{\text{decision}}\bigl(\text{Action}_1,\, \text{Action}_2\bigr).
\end{aligned}
\end{equation}
where stage1 and stage2 denote the approaching and manipulation stages respectively. $Lang$ is the language instruction, $Obs$ is the point cloud, $State$ is the robot proprioception, and \textit{Afford} and $Flow$ are the visual priors.

Although some frameworks \cite{sundaresan2024s, belkhale2023hydra} decompose general robotic manipulation into coarse-grained and fine-grained stages and employ hierarchical imitation learning, their task decoupling is not sufficiently specific. Our task decomposition, in contrast, aligns closely with the nature of object manipulation and we introduce appropriate visual priors for each stage to further enhance model performance.

Moreover, current object manipulation benchmarks \cite{james2020rlbench, chen2025robotwin, wang2025adamanip} define various types of tasks for object manipulation, however, they include few objects and lack annotations for the two visual priors. This hinders evaluation of generalization across instances and different components of the same object, as well as phase-aware decision-making and control. To address this, we leverage the digital assets and part-level annotations from GAPartNet \cite{geng2023gapartnet} and RoboTwin \cite{chen2025robotwin} to construct a comprehensive dataset for training our method and conducting evaluation.

\section{Dual-Actor Policy Modeling}

\begin{figure*}[t]
    \centering
    \includegraphics[width=0.95\textwidth]{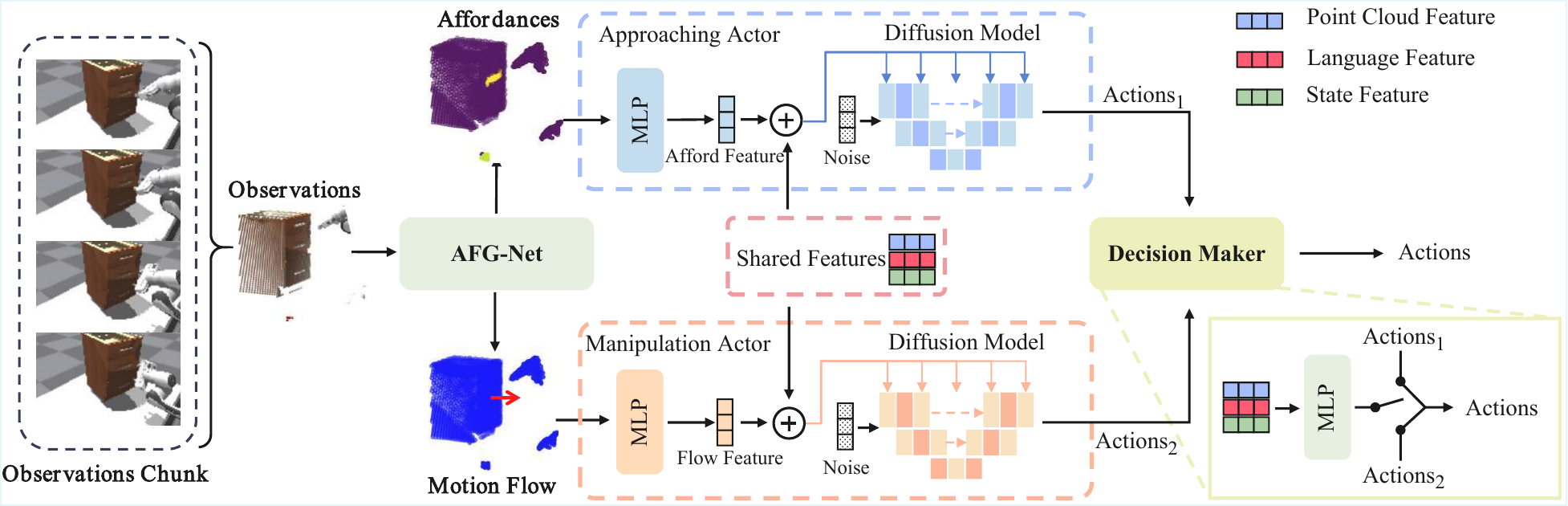}
    \caption{\textbf{Overview of Dual-Actor Policy.} Our framework comprises AFG-Net, an approaching actor, a manipulation actor and a decision maker. Firstly, AFG-Net encodes point cloud observations via a shared encoder, processes language instructions, and predicts affordance and flow through separate decoder heads. Then, the approaching actor uses the predicted affordance to generate actions for the approaching stage, while the manipulation actor uses the predicted motion flow to generate actions for the manipulation stage. Finally, we utilize a decision maker to select which actor produces the actions.}
    \label{fig:overview}
    \vspace{-12pt}
\end{figure*}
In this section, we describe how the Dual-Actor Policy (DAP) addresses the challenges in single-actor models. First, we revisit the motivation behind our method and emphasize the challenges of single-actor models (Sec. \ref{subsec:Motivation}). Following that, we explain how dual actors and heterogeneous visual priors modeling resolve these issues and enhance object manipulation (Sec. \ref{subsec:AFG Net} and Sec. \ref{subsec:Dual-Actor Policy}). Finally, we summarize the overall structure of the Dual-Actor Policy (Sec. \ref{subsec:Overall_Framework}).

\subsection{Challenges in Single-Actor Models}
\label{subsec:Motivation}

Based on the characteristics of this task, object manipulation can be divided into two stages: the approaching and manipulation stages, which are relatively independent of each other. The approaching stage focuses on correctly reaching the specified object components, while the manipulation stage focuses on accurately completing the task based on the properties of the object components. Therefore, it is equally important for the policy to effectively learn the actions for each stage. However, previous works do not fully address this issue, as they rely on a single model to directly learn the entire manipulation process. This approach often results in the model struggling to equally learn the actions for different stages due to the varying duration of each stage and the lack of explicit modeling for learning at different stages.

To validate this, we conduct visualizations using our policy and baselines. As shown in Fig. \ref{fig:Single policy vs. Ours}, single-actor models do not correctly approach the drawer handle and may apply force in an inaccurate direction during manipulation.

\subsection{Dual-Actor Design}
\label{subsec:Dual-Actor Policy}
To address the issue of unequally learning different stages of object manipulation, in this paper, we propose explicitly modeling the learning of the approaching stage and the manipulation stage. Specifically, we assign an independent actor to each stage, where it takes the language instructions $Lang$, the point cloud observations $Obs$ and the state of the robot arm $State$ as inputs and outputs actions for the current stage. As shown in Fig. \ref{fig:Single policy vs. Ours}, by utilizing the dual-actor design, our model outperforms the single-actor baselines. Moreover, as illustrated in Fig. \ref{fig:Single policy vs. Ours}, our model can effectively reach the specified object components and apply the correct force during manipulation. Overall, we consider the dual-actor design to be beneficial for object manipulation. The loss for our dual-actor design is defined as:

\begin{equation}
\begin{aligned}
    \mathcal{L}_{\text{total}} &=
    \begin{cases}
        \gamma \,\mathcal{L}_{\text{stage1}} + (1 - \gamma)\,\mathcal{L}_{\text{stage2}}, & \text{if stage1}, \\[4pt]
        (1 - \gamma)\,\mathcal{L}_{\text{stage1}} + \gamma \,\mathcal{L}_{\text{stage2}}, & \text{if stage2},
    \end{cases} \\[1.75pt]
    \mathcal{L}_{\text{stage1}} &= \mathcal{L}_{\text{action}}\!\left(\theta_{\text{stage1}},\, Lang,\, Obs,\, State\right), \\[1.75pt]
    \mathcal{L}_{\text{stage2}} &= \mathcal{L}_{\text{action}}\!\left(\theta_{\text{stage2}},\, Lang,\, Obs,\, State\right).
\end{aligned}
\end{equation}
where $\mathcal{L}_{\text{action}}$ represents the action generation loss, which will be described in Sec. \ref{subsec:Overall_Framework}. $\theta_{\text{stage1}}$ and $\theta_{\text{stage2}}$ denote the action generators for the approaching stage and the manipulation stage respectively. The parameter $\gamma \in (0.5, 1)$ controls the relative weighting between the primary and secondary phase learning, encouraging each policy to retain a certain level of competence in both phases, thereby improving generalization and failure tolerance.

\begin{figure}[t]
    \centering
    \includegraphics[width=0.95\linewidth]{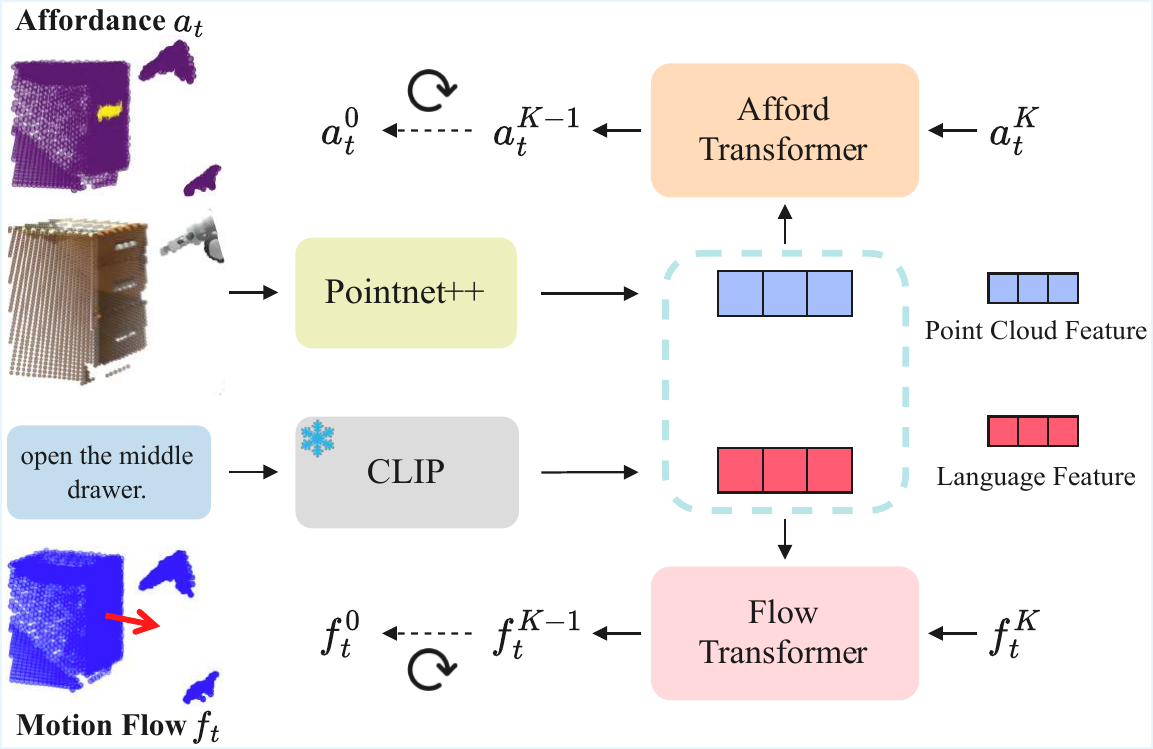}
    \caption{\textbf{Architecture of AFG-Net.} In AFG-Net, we use PointNet++ \cite{qi2017pointnet++} to extract features from point cloud observations, and the CLIP \cite{zhao2023learning} language encoder to obtain features from language instructions. These features are then concatenated and passed separately into the affordance and motion flow prediction heads to obtain heterogeneous visual priors. Here, we adopt the flow matching \cite{lipman2022flow} generative paradigm to predict the visual priors.}
    \label{fig:AFG-Net inference}
    \vspace{-10pt}
\end{figure}

\subsection{Affordance-Flow Generation Network}
\label{subsec:AFG Net}

To further enhance the performance of each actor in its corresponding stage, we introduce heterogeneous visual priors for different stages of object manipulation. As affordances provide effective spatial guidance for contact and indicate where to interact, we use affordances to enhance the approaching stage. Similarly, as motion flow captures the overall movement trends of object parts and informs the expected motion during interaction, we leverage motion flow to promote the manipulation stage.

Thanks to the powerful flow matching \cite{lipman2022flow} generative paradigm, we propose a unified Affordance-Flow Generation Network, termed AFG-Net, for joint affordance and motion flow generation, capturing both geometric and dynamic information within a shared representation. Specifically, as illustrated in Fig. \ref{fig:AFG-Net inference}, AFG-Net employs a shared encoder and two task-specific decoders. The shared encoder leverages the structural and functional correlation between affordance and motion flow on object parts to learn more robust and generalizable joint representations. It is based on PointNet++ \cite{qi2017pointnet++} for processing the point cloud $Obs$ and a pre-trained CLIP model \cite{zhao2023learning} for encoding the language instruction $Lang$. The two decoders are implemented as Perceiver IO Transformers \cite{jaegle2021perceiver}. The training loss of AFG-Net follows the flow matching paradigm \cite{lipman2022flow} and is defined as:
\begin{equation}
\begin{aligned}
    \mathcal{L}_{\text{AFG}} 
        &= \mathcal{L}_{\text{affordance}} + \mathcal{L}_{\text{flow}}, \\[4pt]
    \mathcal{L}_{\text{affordance}} 
        &= \mathbb{E}_{t}\,\bigl\lVert \theta_{\text{AFG}}(\mathbf{a}_t,\, t \mid I) 
        - (\mathbf{a}_1 - \mathbf{a}_0) \bigr\rVert^2, \\[4pt]
    \mathcal{L}_{\text{flow}} 
        &= \mathbb{E}_{t}\,\bigl\lVert \theta_{\text{AFG}}(\mathbf{f}_t,\, t \mid I) 
        - (\mathbf{f}_1 - \mathbf{f}_0) \bigr\rVert^2 .
\end{aligned}
\end{equation}
where $a_0$ and $f_0$ denote samples from a simple base distribution at time $t=0$, and $a_1$ and $f_1$ denote samples from the target complex distribution at time $t=1$. $a_t$ and $f_t$ are defined as the linear interpolation between $a_0$ and $f_0$ as well as $a_1$ and $f_1$ respectively. $\theta_{\text{AFG}}$ denotes the parameters of AFG-Net and $I$ denotes the concatenation of language features, global observation features, point cloud coordinates and encoded Gaussian noise.

\subsection{Overall of Dual-Actor Policy}
\label{subsec:Overall_Framework}

\noindent \textbf{Overall Framework.} Bringing everything together, we develop our Dual-Actor Policy. As shown in Fig. \ref{fig:overview}, we first utilize observations (in this paper, we use point clouds) and the language instruction as inputs, and output the affordance and motion flow of the specific component of the target object using AFG-Net. Then, we combine the predicted affordance and motion flow with the proposed dual-actor design, where we incorporate affordance with the approaching actor to enhance the reaching process, while combining motion flow with the manipulation actor to promote execution. Finally, we propose a Decision Maker that autonomously selects which actor to call for action generation based on the current scene observations. Benefiting from our dual-actor design and heterogeneous visual priors, our method fully considers the characteristics of different stages of object manipulation, effectively improving the performance. 

\noindent \textbf{Loss Design.} Our Dual-Actor Policy uses DDIM \cite{ddim} as the action generator and the ground truth is represented by the sample. The loss $\mathcal{L}_{\text{actor}}$ is defined as:
\begin{equation}
\mathcal{L}_{\text{actor}}(\theta, I) = \mathbb{E}_{t, z \sim \mathcal{D}(x)} \left\| z - \theta(z_t, t \mid I) \right\|^2,
\end{equation}
where $z \sim \mathcal{D}(x)$ denotes the action distribution from expert demonstrations, $I$ represents the conditions and $\theta$ denotes the DDIM-based action generator.

Following that, combining everything together, the loss of our Dual-Actor Policy can be updated as:
\begin{equation}
\begin{aligned}
    \mathcal{L}_{\text{total}} &=
    \begin{cases}
        \gamma \,\mathcal{L}_{\text{stage1}} + (1 - \gamma)\,\mathcal{L}_{\text{stage2}}, & \text{if stage 1}, \\[1.75pt]
        (1 - \gamma)\,\mathcal{L}_{\text{stage1}} + \gamma \,\mathcal{L}_{\text{stage2}}, & \text{if stage 2},
    \end{cases} \\[1.75pt]
    \mathcal{L}_{\text{stage1}} &= \mathcal{L}_{\text{action}}\!\left(\theta_{\text{stage1}},\, Lang,\, Obs,\, State,\, \textit{Afford}\right), \\[1.75pt]
    \mathcal{L}_{\text{stage2}} &= \mathcal{L}_{\text{action}}\!\left(\theta_{\text{stage2}},\, Lang,\, Obs,\, State,\, Flow\right).
\end{aligned}
\end{equation}
where \textit{Afford} and $Flow$ represent the predicted affordance and flow generated by AFG-Net.

\noindent \textbf{Implementation Details.} We first train AFG-Net and use it to augment the dataset with imperfect affordance and motion flow, ensuring consistency between training and inference. During the training of DAP, we freeze the AFG-Net and train the two actors and the decision maker. In our dataset, we use 6D end-effector poses for proprioception and prediction, while in RoboTwin and real-world settings, we use and predict joint angles. In the loss function, $\gamma$ is set to 0.75. Our policy is implemented using PyTorch and trained on a single NVIDIA RTX 4090 GPU for 2000 epochs with the AdamW optimizer and a batch size of 80.

\section{Dual-Prior Object Manipulation Dataset}
\label{subsec:dataset preparation}
\subsection{Motivation of Dual-Prior Object Manipulation Dataset} 
Existing object manipulation benchmarks contain few objects and lack annotations for the two stages. More importantly, our Dual-Actor Policy requires heterogeneous visual priors for training. To support both training and evaluation, we propose a large-scale dataset, the Dual-Prior Object Manipulation Dataset, based on the IsaacGym \cite{makoviychuk2021isaac} platform and the GAPartNet \cite{geng2023gapartnet} dataset. It contains 12 object categories, 563 objects and a total of 31,501 expert demonstrations. Each expert demonstration includes RGB-D images, language instructions, affordance and flow annotations, segmentation maps, object and camera poses, and stage labels.

\subsection{Affordance Generation}
Since affordance reflects the functional relevance between regions and manipulation tasks, we assume that points closer to the task-relevant component should have higher values. To model this property, we compute the shortest Euclidean distance from each point $\mathbf{p}$ in the scene point cloud to the point cloud of the key functional component (e.g., the handle of a drawer). Let $\mathcal{P}_{\text{key}} = \{\mathbf{k}_1, \mathbf{k}_2, \dots, \mathbf{k}_N\}$ denote the set of 3D points representing the key functional part. The distance from $\mathbf{p}$ to this component is then defined as:
\begin{equation}
d(\mathbf{p}) = \min_{\mathbf{k} \in \mathcal{P}_{\text{key}}} \|\mathbf{p} - \mathbf{k}\|_2,
\end{equation}
where $d(\mathbf{p})$ captures the closest proximity between $\mathbf{p}$ and any point on the key functional part. Instead of using the raw inverse distance, we pass the distance through a modulated sigmoid function to obtain a smooth and bounded affordance map within $[-1, 1]$. The value is computed as:
\begin{equation}
\mathcal{A}(\mathbf{p}) = 3 - 4( \sigma(\alpha \cdot d(\mathbf{p}))),
\end{equation}
where $\sigma(\cdot)$ denotes the sigmoid function and $\alpha > 0$ is a scaling factor that controls the sharpness of the transition.

\subsection{Motion Flow Generation}
For the generation of motion flow, we follow the definition paradigm of Flowbot3D \cite{eisner2022flowbot3d} and compute the instantaneous velocities of objects based on the differences in joint angles or rigid poses between consecutive frames.

Given a point $\mathbf{p}_i \in \mathbb{R}^3$ on a moving part in the current frame, let $J_i$ and $J_{i+1}$ denote the joint parameters at times $t$ and $t+\Delta t$ respectively. We define a forward kinematics mapping $T(J): \mathbb{R}^3 \to \mathbb{R}^3$ that computes and transforms the 3D point coordinates according to the given joint state. The corresponding position of $\mathbf{p}_i$ in the next frame, as well as the motion flow $\mathbf{F}_i$, are then predicted as:
\begin{equation}
\begin{aligned}
    \mathbf{p}_{i+1} &= T(J_{i+1}) \bigl( T(J_i)^{-1} (\mathbf{p}_i) \bigr), \\[1.75pt]
    \mathbf{F}_i &= \frac{\mathbf{p}_{i+1} - \mathbf{p}_i}{\Delta t}.
\end{aligned}
\end{equation}

We then apply normalization to ensure scale invariance. (1) For prismatic joints (e.g., drawers) and unarticulated objects (e.g., bottles), all moving points have the same velocity magnitude. Therefore, we rescale $\mathbf{F}_i$ such that $|\mathbf{F}_i| = 1$. (2) For revolute joints (e.g., doors), we normalize $\mathbf{F}_i$ by the maximum expected speed $v_{\max}$, which is defined as:
\begin{equation}
\begin{aligned}
    v_{\max} &= |J_{i+1} - J_i| \cdot r_{\max}, \\
    \mathbf{F}_i &= \frac{\mathbf{F}_i}{v_{\max}}.
\end{aligned}
\end{equation}

\subsection{Task Phase Annotation}
Considering the characteristics of object manipulation, which can be divided into two stages: the approaching stage and the manipulation stage, we adopt the open or closed state of the gripper as a heuristic to distinguish between these stages. Specifically, when the gripper transitions from the open state to the closed state, the approaching stage ends and the manipulation stage begins. Conversely, when the gripper transitions from the closed state to the open state, the manipulation stage ends, marking the completion of the current object manipulation task.

\section{Experiments}
\label{sec:Experiments}

\begin{figure}
    \centering
    \hfill
	\includegraphics[width=0.95\linewidth]{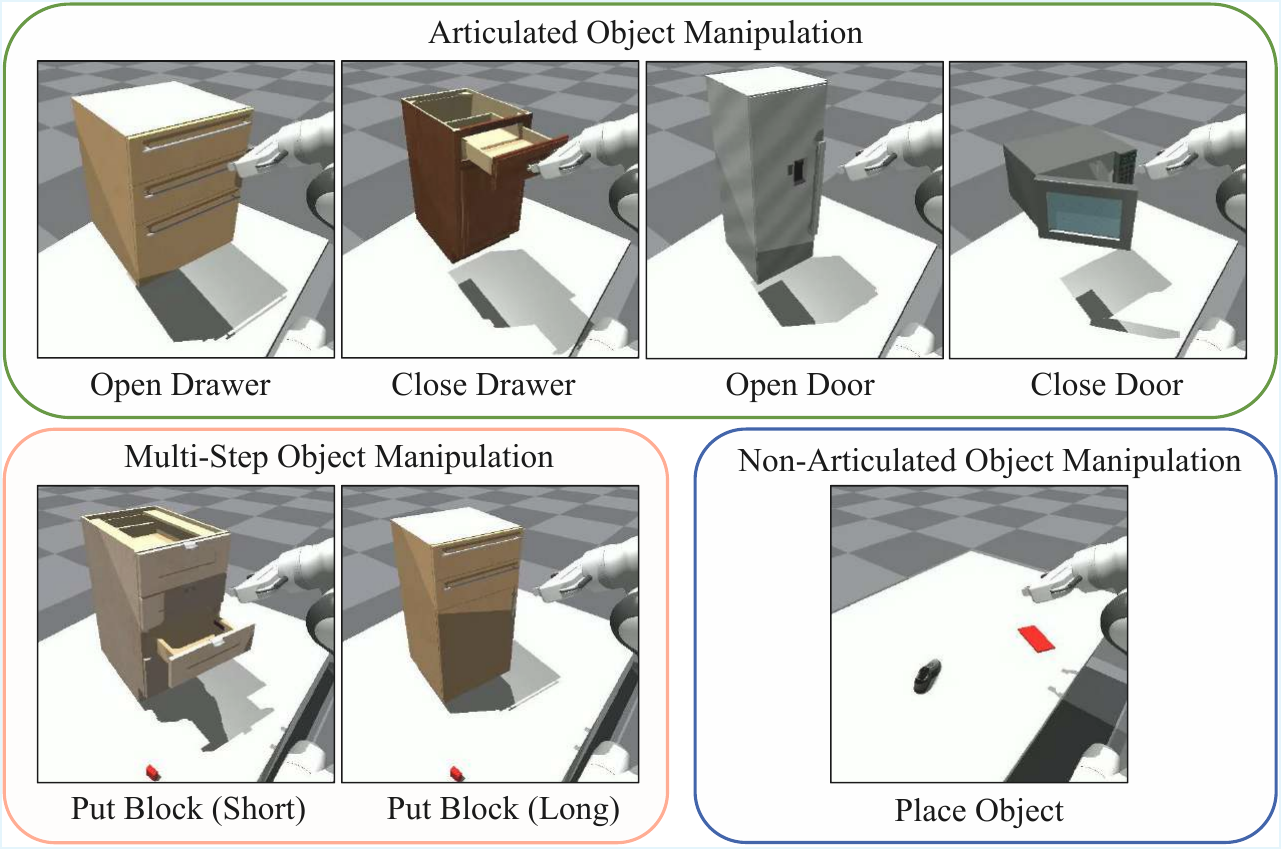}
	\caption{\textbf{Seven tasks in our dataset.} Our dataset covers three categories: articulated object manipulation, multi-step object manipulation and non-articulated object manipulation. Zoom in for a better view.}
	\label{fig:tasks}
    \vspace{-10pt}
\end{figure}

\newcommand{\best}[1]{\textbf{#1}}     
\newcommand{\second}[1]{\ul{#1}}       
\definecolor{darkgreen}{rgb}{0, 0.3, 0}

\begin{table*}[t]
\centering
\setlength{\tabcolsep}{4pt}
\resizebox{0.95\textwidth}{!}{
\begin{tabular}{l | c c c c c c c}
\toprule
 
& \textbf{ACT} \cite{zhao2023learning} 
& \textbf{DP} \cite{chi2023diffusion} 
& \textbf{DP3} \cite{Ze2024DP3} 
& \textbf{AffordDP} \cite{wu2025afforddp} 
& \textbf{RDT} \cite{liu2024rdt1b} 
& $\boldsymbol{\pi_0}$ \cite{black2024pi_0} 
& \textbf{Ours} \\
\toprule
Open Drawer     & \phantom{0}0.1 / \phantom{0}0.0 & \phantom{0}0.4 / \phantom{0}0.2 & 18.2 / 13.0 & \phantom{0}1.2 / \phantom{0}1.9 & \second{33.1} / \second{23.5} & 20.8 / 15.1 & \best{48.5 / 36.1} \\
Close Drawer    & 28.7 / 22.7 & \phantom{0}9.3 / \phantom{0}8.5 & 50.3 / 41.6 & 46.5 / 43.8 & 65.4 / 55.3 & \best{80.3} / \best{65.9} & \second{74.5} / \second{58.7} \\
Open Door       & \phantom{0}0.3 / \phantom{0}0.4 & \phantom{0}0.4 / \phantom{0}0.0 & \phantom{0}6.4 / \phantom{0}3.0 & \phantom{0}0.7 / \phantom{0}0.2 & 19.0 / 13.1 & \second{21.1} / \second{19.9} & \best{28.8 / 25.4} \\
Close Door      & \phantom{0}8.3 / \phantom{0}9.2 & \phantom{0}1.7 / \phantom{0}0.9 & 47.0 / 44.8 & 42.3 / 39.3 & \best{65.9} / \best{65.7} & 49.9 / 47.1 & \second{58.1} / \second{58.4} \\
Put Block (Short) & \phantom{0}0.0 / \phantom{0}0.0 & \phantom{0}0.0 / \phantom{0}0.0 & 16.3 / 13.9 & \phantom{0}1.6 / \phantom{0}1.6 & \phantom{0}8.5 / \phantom{0}6.8 & \second{21.3} / \second{18.1} & \best{34.1 / 34.4} \\
Put Block (Long) & \phantom{0}0.0 / \phantom{0}0.0 & \phantom{0}0.0 / \phantom{0}0.0 & \phantom{0}3.7 / \phantom{0}2.1 & \phantom{0}0.0 / \phantom{0}0.0 & \phantom{0}0.8 / \phantom{0}1.1 & \second{14.9} / \second{11.8} & \best{15.2 / 14.5} \\
Place Object    & \phantom{0}0.0 / \phantom{0}0.0 & \phantom{0}0.0 / \phantom{0}0.0 & \phantom{0}2.4 / \phantom{0}1.5 & \phantom{0}0.0 / \phantom{0}0.0 & \phantom{0}2.4 / \phantom{0}2.6 & \best{28.7 / 20.2} & \second{15.1} / \second{12.1} \\
\midrule
\rowcolor{gray!20}
Average         & \phantom{0}5.3 / \phantom{0}4.6 & \phantom{0}1.7 / \phantom{0}1.4 & 20.6 / 17.1 & 13.2 / 12.4 & 27.9 / 24.0 & \second{33.8} / \second{28.5} & \best{39.2 \scriptsize{(\textcolor{darkgreen} {$\uparrow$ 5.4})} / 34.2 \scriptsize{(\textcolor{darkgreen} {$\uparrow$ 5.7})}} \\
\bottomrule
\end{tabular}
}
\caption{\textbf{Performance comparison on our dataset.} All values are reported in percentage (\%). In this table, the left value in each cell represents the success rate on the seen object set and the right value denotes the success rate on the unseen object set. The best result for each task is highlighted in \textbf{bold} and the second-best result is \second{underlined}.}
\label{table:Main}
\end{table*}
\begin{table*}[t]
\centering
\setlength{\tabcolsep}{4pt} 
\resizebox{0.725\textwidth}{!}{
\begin{tabular}{l | c c c c c c}
\toprule
 
& \textbf{ACT} \cite{zhao2023learning} 
& \textbf{DP} \cite{chi2023diffusion} 
& \textbf{DP3} \cite{Ze2024DP3} 
& \textbf{RDT} \cite{liu2024rdt1b} 
& $\boldsymbol{\pi_0}$ \cite{black2024pi_0} 
& \textbf{Ours} \\
\midrule
Pick Diverse Bottles & \phantom{0}3.0 & \phantom{0}6.0 & \second{52.0} &  \phantom{0}2.0 & 27.0 & \best{68.0} \\
Put Object Cabinet   & \phantom{0}0.0 & 42.0 & \second{72.0} &  33.0 & 68.0 & \best{78.0} \\
Place Dual Shoes     & \phantom{0}0.0 & \phantom{0}8.0 & 13.0 &  \phantom{0}4.0 & \second{15.0} & \best{18.0} \\
Place Shoe           & \phantom{0}1.0 & 36.0 & \second{58.0} &  35.0 & 28.0 & \best{74.0} \\
Open Laptop          & 32.0 & 46.0 & 82.0 &  59.0 & \second{85.0} & \best{88.0} \\
Open Microwave       & \phantom{0}0.0 & \phantom{0}5.0 & 61.0 &  37.0 & \second{80.0} & \best{100.0\phantom{0}} \\
\midrule
\rowcolor{gray!20}
Average      & 6.0 & 23.8 & \second{56.3} &  28.3 & 50.5 & \best{71.0 (\textcolor{darkgreen} {$\uparrow$ 14.7})} \\
\bottomrule
\end{tabular}
}
\caption{\textbf{Performance comparison on RoboTwin 2.0.} All values are reported in percentage (\%). The best result for each task is highlighted in \textbf{bold} and the second-best result is \second{underlined}. }
\label{table:robotwin}
\vspace{-5pt}
\end{table*}

\subsection{Experimental Setup}
\label{sec:Experimental Setup}
\noindent \textbf{Simulation Benchmark.}
We validate our policy across two distinct platforms to ensure a comprehensive evaluation: our Dual-Prior Object Manipulation Dataset and the RoboTwin 2.0 benchmark \cite{chen2025robotwin}. It should be noted that we select several dual-arm object manipulation tasks from the RoboTwin 2.0 to verify that our policy remains effective on dual-arm robots.

\noindent \textbf{Tasks in Our Dataset.}
We construct seven object manipulation tasks as shown in Fig. \ref{fig:tasks}:
(1) Articulated Object Manipulation: opening and closing doors and drawers of articulated objects under language instructions, requiring precise interaction with hinged parts.
(2) Non-Articulated Object Manipulation: We design a task involving fine-grained manipulation of non-articulated objects under geometric constraints, i.e., placing a randomly positioned object onto a mat while aligning its orientation with the mat.
(3) Multi-Step Object Manipulation: We sequence multiple object manipulation tasks to form a relatively long-horizon and multi-step manipulation task. The Put Block (Short) includes two phases: placing the block into a pre-opened drawer and then closing the drawer. The Put Block (Long) includes three phases: opening the specified drawer based on instructions, placing the block inside and closing it.

\noindent \textbf{Tasks in RoboTwin 2.0 benchmark.}
We further verify our method on dual-arm robots using the RoboTwin 2.0 benchmark, selecting six tasks to assess the effectiveness of our method in bimanual object manipulation. These include 
(1) Dual-arm non-articulated object manipulation tasks: Pick Diverse Bottles and Place Dual Shoes. 
(2) Dual-arm articulated object manipulation tasks: Put Object Cabinet. 
(3) Single-arm non-articulated object manipulation tasks: Place Shoe. and 
(4) Single-arm articulated object manipulation tasks: Open Laptop and Open Microwave. 
Notably, although some tasks involve only one arm, the robot needs to autonomously decide which arm to use based on the location of the object.
\noindent \textbf{Baselines.} 
We evaluate our method against several representative baselines: 
(1) Diffusion-based policies: Diffusion Policy (DP) \cite{chi2023diffusion}, 3D Diffusion Policy (DP3) \cite{Ze2024DP3} and AffordDP \cite{wu2025afforddp}, which extends DP3 by incorporating affordance as an explicit prior.
(2) Regression-based model: ACT \cite{zhao2023learning}, a model that takes multi-view images as input and directly regresses actions using an action chunking mechanism.
(3) Vision-Language-Action models: RDT \cite{liu2024rdt1b} and $\pi_0$ \cite{black2024pi_0}, representing large-scale, general-purpose VLA models.

\subsection{Evaluation in Simulations}
\label{sec:Simulation Performance}

\noindent \textbf{Quantitative Results.} 
In our dataset, we collect RGB-D images from three viewpoints, each with a resolution of $128 \times 128$. We also construct a fused 3D point cloud by merging observations from the three cameras and then apply farthest point sampling to downsample the point cloud to 4,096 points. The experimental results are reported in Tab. \ref{table:Main}. It is evident that: (1) Our policy achieves SOTA performance on most tasks, significantly outperforming non-VLA methods. (2) In the evaluation of generalization on unseen objects, our policy outperforms all baselines on average, demonstrating that the combination of heterogeneous visual priors and the dual-actor design enhances generalization to unseen objects. (3) For multi-stage tasks, DAP also achieves SOTA, showing its effectiveness in handling detailed decision making and control across multiple phases.

In the RoboTwin 2.0 benchmark, we collect 50 expert demonstrations and use point clouds downsampled to 1,024 points for consistency with the original setup. As shown in Tab. \ref{table:robotwin}, our method continues to achieve strong performance and SOTA results on several tasks, further demonstrating its effectiveness and generalizability to dual-arm robots.

\begin{figure*}[t]
    \centering
    \includegraphics[width=0.9\textwidth]{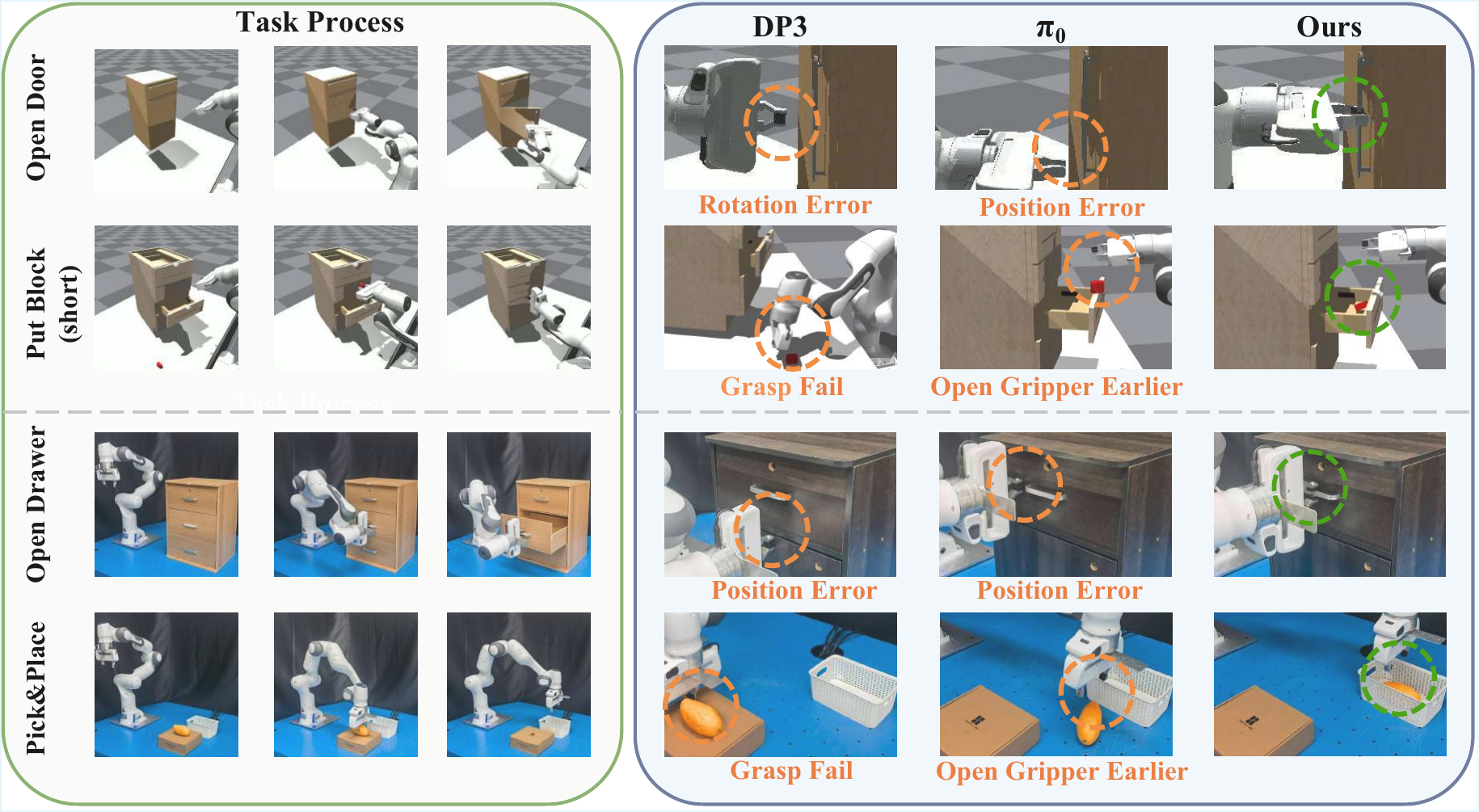}
    \caption{\textbf{Visualization of task execution processes and results from different methods.} The left panel illustrates the task execution processes for two simulated tasks from our dataset and two real-world tasks. The right panel presents a qualitative comparison of the execution trajectories produced by DP3~\cite{Ze2024DP3}, $\pi_0$~\cite{black2024pi_0}, and our method. Zoom in for the best view.}
    \label{fig:visualization}
    \vspace{-5pt}
\end{figure*}

\noindent \textbf{Qualitative Results.}
As shown in Fig. \ref{fig:visualization}, DP3 \cite{Ze2024DP3} frequently suffers from grasping failures due to inaccurate perception of critical object components, preventing it from completing subsequent manipulation steps. This indicates a limited ability to localize key interaction points under complex visual conditions. $\pi_0$ \cite{black2024pi_0} often exhibits drifting in pose estimation during task execution, causing misalignment between the gripper and the target component and leading to failure. In contrast, our method achieves more robust and precise control thanks to its dual-actor architecture and heterogeneous visual priors. These designs enable accurate reasoning about both the approach to manipulating functional parts and the identification of the correct manipulation location.

\subsection{Ablation Study}
\label{sec:Ablation Study}
\newcommand{\checkmarkblack}{\textcolor{black}{\checkmark}}

\begin{table}[t]
    \centering
    {
    \resizebox{0.95\columnwidth}{!}{
    \begin{tabular}{c c c | c}
        \toprule
        Affordance & Motion Flow & Dual-Actor Design & Average \\
        \midrule
        &                   &                   & 18.7 \phantom{(\textcolor{darkgreen}{$\uparrow$ \phantom{0}7.9})} \\
        \checkmarkblack &                   &                   & 26.6 (\textcolor{darkgreen}{$\uparrow$ \phantom{0}7.9}) \\
                        & \checkmarkblack   &                   & 28.2 (\textcolor{darkgreen}{$\uparrow$ \phantom{0}9.5}) \\
                        &                   & \checkmarkblack   & 27.7 (\textcolor{darkgreen}{$\uparrow$ \phantom{0}9.0}) \\          
        \checkmarkblack & \checkmarkblack   &                   & 30.7 (\textcolor{darkgreen}{$\uparrow$ 12.0}) \\
        \checkmarkblack & \checkmarkblack   & \checkmarkblack   & 36.7 (\textcolor{darkgreen}{$\uparrow$ 18.0}) \\
        \bottomrule
    \end{tabular}
        }
    }
    \caption{\textbf{Ablation study.} We evaluate different ablated variants on our dataset and report their average success rates. The first row corresponds to the baseline method, while the last row represents our full model.}
    \label{tab:ablation}
    \vspace{-12pt}
\end{table}
To assess the effectiveness of the core components, the AFG-Net and the Dual-Actor design in DAP, we conduct ablation studies. The results are summarized in Tab. \ref{tab:ablation}. First, we adopt a single-policy model as the baseline, achieving a success rate of 18.7\%. Incorporating individual visual priors yields noticeable improvements: augmenting the baseline with Affordance or Motion Flow priors increases the success rate to 26.6\% and 28.2\% respectively. Next, employing the Dual-Actor design alone improves performance to 27.7\%, demonstrating the advantage of this modular structure. Moreover, combining both visual priors further enhances performance, raising the success rate to 30.7\%, indicating the effectiveness of the complementary priors. Finally, our full Dual-Actor Policy, which combines the heterogeneous visual priors with the Dual-Actor design, achieves the highest success rate of 36.7\%, highlighting the effectiveness of proposed modular architectures for object manipulation.

\subsection{Evaluation in Real-World Scenarios}
\label{sec:Real-world Performance}

\begin{table}[t]
    \centering
    \setlength{\tabcolsep}{2pt}
    \begin{tabular}{l *{8}{c}}
        \toprule

        & \multicolumn{2}{c}{Open Drawer} 
        & \multicolumn{2}{c}{Close Drawer} 
        & \multicolumn{2}{c}{Close Door} 
        & \multicolumn{2}{c}{Pick-and-Place} \\
        \cmidrule(lr){2-3} \cmidrule(lr){4-5} \cmidrule(lr){6-7} \cmidrule(l){8-9}
        & Seen & Unseen
        & Seen & Unseen 
        & Seen & Unseen 
        & Seen & Unseen\\
        \midrule
        DP \cite{chi2023diffusion}   & 10.0 & 5.0   & 30.0  & 25.0 & 60.0 & 40.0 & 26.7 & 6.7 \\
        DP3 \cite{Ze2024DP3}         & 20.0 & 5.0   & 30.0  & 20.0 & \ul{70.0}& 55.0  & 20.0 & 6.7 \\
        RDT \cite{liu2024rdt1b}      & 40.0 & 20.0  & 60.0  & 20.0 & \textbf{80.0}& \ul{70.0}  & 40.0 & 30.0 \\
        ${\pi_0}$ \cite{black2024pi_0}     & \ul{50.0} & \ul{30.0}  & \ul{70.0}  & \ul{35.0} & \textbf{80.0} & \textbf{75.0}  & \ul{53.3} & \ul{33.3} \\
        \rowcolor{gray!20}
        Ours                        & \textbf{70.0} & \textbf{35.0}  & \textbf{80.0}  & \textbf{60.0} & \textbf{80.0}& \textbf{75.0}  & \textbf{63.3} & \textbf{46.7} \\
        \bottomrule
    \end{tabular}
    \caption{\textbf{Real-world experiments.} The best result is shown in \textbf{bold} and the second-best result is \ul{underlined}.}
    \label{table:Real}
    \vspace{-10pt}
\end{table}

\begin{figure}[t]
    \centering
    \includegraphics[width=0.8\linewidth]{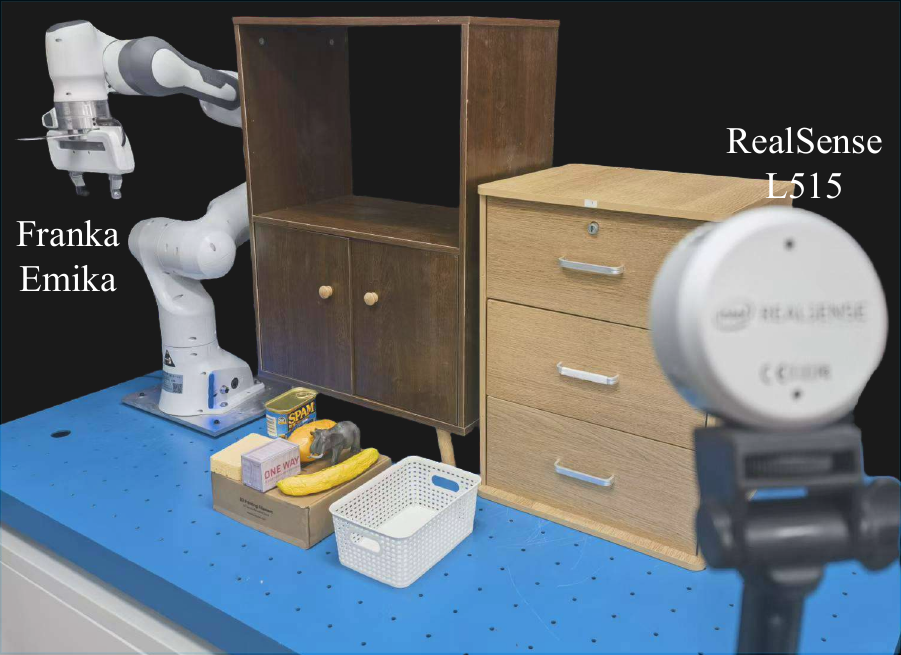}
    \caption{\textbf{Real-world experimental settings.} We use a Franka Emika arm and evaluate diverse everyday objects, with a RealSense L515 camera providing visual perception.} 
    \label{fig:real_world_setting}
    \vspace{-12pt}
\end{figure}

\noindent \textbf{Settings.} To verify the practical capability of our method, we conduct real-world experiments on four representative tasks: (1) Open Drawer, (2) Close Drawer, (3) Close Door and (4) Pick and Place Objects. The real-world experimental settings are illustrated in Fig. \ref{fig:real_world_setting}.

For Tasks 1–3, we collect 30 demonstrations on a single seen object, while reserving two additional objects as unseen for evaluation. For the Pick and Place Objects task, the robot is required to grasp an object from the table and place it into a basket. We collect 15 demonstrations for each of three seen objects (Sponge, Banana, Mango), resulting in a total of 45 episodes. Three other unseen objects (Tiny Pink Container, Canned Luncheon Meat, Elephant Toy) are used during evaluation to assess generalization.

\noindent \textbf{Experimental Results.} 
We evaluate the performance of our method in real-world scenarios by repeating each task-object pair 10 times and reporting the corresponding success rate. As illustrated in Tab. \ref{table:Real}, for articulated object manipulation tasks (Open Drawer, Close Drawer and Close Door), our method achieves SOTA results. For the non-articulated object manipulation task Pick and Place Objects, which requires precise grasping and placement, our method achieves a success rate of 63.3\% on seen objects and 46.7\% on unseen objects, surpassing all compared methods. This demonstrates stronger generalization compared with diffusion-based and vision-language-action baselines.

Furthermore, we provide qualitative results in Fig. \ref{fig:visualization}. It can be seen that baseline methods such as DP3 and $\pi_0$ often fail during the approaching stage due to the lack of explicit affordance priors, resulting in incorrect localization. In addition, the absence of motion flow priors sometimes leads to placement errors. In contrast, our method effectively mitigates these issues, achieving more reliable task execution and higher success rates.

\subsection{Comparison of Efficiency and Performance}
\label{sec:Method Efficiency}

\begin{table}[t]
    \centering
    \vspace{0.5em}
    \renewcommand{\arraystretch}{1.1}
    \begin{tabular}{l | c c}
        \toprule
        {Method} & {Frequency (Hz)} & {Success Rate (\%)} \\
        \midrule
        DP3 \cite{Ze2024DP3}       & \textbf{8.40} & 18.70    \\
        RDT \cite{liu2024rdt1b}       & 2.75          & 25.94   \\
        ${\pi_0}$ \cite{black2024pi_0}       & 4.54          & \ul{31.07} \\
        AffordDP \cite{wu2025afforddp}  & 2.27          & 9.33    \\
        Ours      & \ul{5.65}     & \textbf{36.70} \\
        \bottomrule
    \end{tabular}
    \caption{\textbf{Comparison of efficiency and performance.} The best result for each task is highlighted in \textbf{bold} and the second-best result is \ul{underlined}.}
    \label{tab:efficiency}
    \vspace{-10pt}
\end{table}

To evaluate efficiency and performance, we measure the inference speed and performance on a single NVIDIA RTX 4090. As reported in Tab. \ref{tab:efficiency}, although our method is slower than DP3 \cite{Ze2024DP3} in inference, it significantly outperforms DP3 in terms of performance. Beyond DP3, our method achieves both higher efficiency and stronger performance compared with other object manipulation methods such as AffordDP \cite{wu2025afforddp} and VLA models like RDT \cite{liu2024rdt1b} and $\pi_0$ \cite{black2024pi_0}.

\section{Conclusion}
\label{sec:Conclusion}

In summary, in this paper, we tackle the problem of how to leverage the characteristics of object manipulation, which can be divided into approaching and manipulation stages, to improve manipulation performance. Previous methods mostly handle this with a single policy, which ignores the staged nature of object manipulation. To address this, we propose a novel Dual-Actor Policy, termed DAP, which explicitly considers different stages and uses heterogeneous visual priors to enhance each stage. This policy includes an affordance-based actor to identify the functional part and a motion flow-based actor to capture the movement of the component. In addition, it includes a decision maker to determine the current stage. To support training with heterogeneous visual priors, we build a Dual-Prior Object Manipulation Dataset with two visual priors and seven tasks. We evaluate our method on our dataset, the RoboTwin benchmark and real-world scenarios, which demonstrates the effectiveness and efficiency of our policy.

\IEEEtriggeratref{39}





\bibliographystyle{IEEEtran}
\bibliography{references}


\end{document}